**Evaluation of GPT-based large language generative AI models as study aids for the national licensure examination for registered dietitians in Japan.**

Yuta Nagamori [a,†], Mikoto Kosai [a,†], Yuji Kawai [a,†], Haruka Marumo [a], Misaki Shibuya [a], Tatsuya Negishi [a], Masaki Imanishi [b], Yasumasa Ikeda [c], Koichiro Tsuchiya [b], Asuka Sawai [d], and Licht Miyamoto [a,*]

[a] *Laboratory of Physiology, Pharmacology and Food Science, Department of Nutrition and Life Science, Faculty of Health and Medical Sciences, Kanagawa Institute of Technology, 1030, Shimo-Ogino, Atsugi-shi, Kanagawa 243-0292, Japan*

[b] *Department of Medical Pharmacology, Institute of Biomedical Sciences, Graduate School of Tokushima University. 1-78-1, Sho-machi, Tokushima 770-8505, Japan*

[c] *Department of Pharmacology, Institute of Biomedical Sciences, Graduate School of Tokushima University. 3-18-15, Kuramoto-cho, Tokushima 770-8505, Japan*

[d] *Department of Nutrition and Life Science, Faculty of Health and Medical Sciences, Kanagawa Institute of Technology, 1030, Shimo-Ogino, Atsugi-shi, Kanagawa 243-0292, Japan*

Running title;
LLMs as Study Aids for Dietitian License Exam

*: Address correspondence and reprint requests to Licht Miyamoto, 1030, Shimo-Ogino, Atsugi-shi, Kanagawa 243-0292, Japan. Tel/fax; +81-46-291-3345, Email: licht_corresp2011[at]yahoo.co.jp (L.M.)

†: Y.N., M.K., and Y.K. equally contributed to the study.

**Acknowledgments**
This study was partly supported by Information Education Research Center, Kanagawa Institute of Technology.
We thank Drs. Hiroshi Tanaka, and Kosuke Takano for their valuable advice and help.

*Yuta Nagamori, Mikoto Kosai, Yuji Kawai, and Licht Miyamoto mainly performed data acquisition, and prepared figures. Licht Miyamoto conceived, designed, and conducted the study, and wrote the manuscript. Yuta Nagamori and Mikoto Kosai also contributed to the study design. The others contributed to the data acquisition and/or interpretation.*

**declaration of interest statement**

The authors declare no conflict of interest related to the current study.

**Abstract**

Introduction:
Generative artificial intelligence (AI) based on large language models (LLMs), such as ChatGPT, has demonstrated remarkable progress across various professional fields, including medicine and education. However, their performance in nutritional education, especially in Japanese national licensure examination for registered dietitians, remains underexplored. This study aimed to evaluate the potential of current LLM-based generative AI models as study aids for nutrition students.

Methods:
Questions from the Japanese national examination for registered dietitians were used as prompts for ChatGPT and three Bing models (Precise, Creative, Balanced), based on GPT-3.5 and GPT-4. Each question was entered into independent sessions, and model responses were analyzed for accuracy, consistency, and response time. Additional prompt engineering, including role assignment, was tested to assess potential performance improvements.

Results:
Bing-Precise (66.2%) and Bing-Creative (61.4%) surpassed the passing threshold (60%), while Bing-Balanced (43.3%) and ChatGPT (42.8%) did not. Bing-Precise and Bing-Creative generally outperformed others across subject fields except Nutrition Education, where all models underperformed. None of the models consistently provided the same correct responses across repeated attempts, highlighting limitations in answer stability. ChatGPT showed greater consistency in response patterns but lower accuracy. Prompt engineering had minimal effect, except for modest improvement when correct answers and explanations were explicitly provided.

Discussion:
While some generative AI models marginally exceeded the passing threshold, overall accuracy and answer consistency remained suboptimal. Moreover, all the models demonstrated notable limitations in answer consistency and robustness. Further advancements are needed to ensure reliable and stable AI-based study aids for dietitian licensure preparation.

---

**Introduction**

One of the recent advances in natural language processing has been the development and widespread adoption of large language models (LLMs). There has been a significant surge in the utilization of the LLM-based generative artificial intelligence (AI) technologies, such as ChatGPT (OpenAI, San Francisco, CA) and Bing (Microsoft Corp., Redmond, WA), leading to a notable change in the landscape of AI-driven applications. This proliferation can be attributed to the rapid advancements in deep learning algorithms, coupled with the growing availability of large-scale datasets for training. As a result, LLM-based generative AI models have made unprecedented progress, enabling them to effectively tackle complex tasks across various domains, including medical diagnostics, and educational assistance (Vrdoljak et al., 2025; Xu et al., 2024; Zhou et al., 2024). The widespread adoption of these technologies underscores their transformative potential as well as the need for comprehensive evaluation and validation in specialized fields such as healthcare and education.

Powered by the deep learning techniques and trained on vast amounts of text data, ChatGPT represents a paradigm shift in how machines generate text. With its ability to generate coherent and contextually relevant responses to a wide range of prompts, ChatGPT has quickly become a cornerstone in various AI applications, ranging from virtual assistants to automated content generation. The success of ChatGPT has spurred the emergence of various AI competitors such as Bing (rebranded as Copilot by Microsoft) and Bard (rebranded as Gemini by Google LLC, Mountain View, CA). Their versatility and adaptability make them valuable tools not only for researchers and developers but also for educators and professionals across various industries. As such, understanding the capabilities and limitations of LLM-based generative AI models is essential for harnessing their full potential in the real world.

In Japan, the profession of registered dietitian, which represents a more advanced qualification than dietitian, holds a significant role in promoting public health and nutrition education for the public and patients. Governed by strict regulatory standards and professional guidelines, the pathway to becoming a registered dietitian involves rigorous academic training, practical experience, and successful completion of the national licensure examination administered by the Ministry of Health, Labor and Welfare. Registered dietitians are trained to provide evidence-based nutrition counseling, develop personalized dietary plans, and contribute to the prevention and management of various health conditions, including obesity, diabetes, and cardiovascular diseases. With an increasing emphasis on preventive healthcare and dietary management, the demand for qualified registered dietitians continues to grow, highlighting the importance of robust training programs and ongoing professional development in this field.

The national registered dietitian licensure examination in Japan serves as a pivotal milestone in the educational course to becoming a certified registered dietitian. This rigorous examination assesses candidates' knowledge, skills, and competencies across various domains like nutrition, food science, gastronomy, physiology and public health. The examination comprises 200 multiple-choice questions including case studies of practical assessments, designed to evaluate candidates' ability to apply theoretical knowledge to real-world scenarios. The typical passing criterion is a correct answer rate of 60%. The examination is divided into two parts, each consisting of five domains: social nutrition, structure and function of the human body and pathogenesis of diseases, food science in relation to our health, basic and applied nutrition, nutrition education, clinical nutrition, public nutrition, food service management, and their practical applications. Then, these two parts collectively cover various aspects of nutrition and dietetics, ensuring candidates' comprehensive understanding and readiness to practice as registered dietitians. There are no cutoff scores set for the individual domains.

As aforementioned, the LLM-based generative AI models have been proposed to be applied for many specialty fields including medicine and education **(Xu et al., 2024)**. We speculated that the LLM-based generative AI models could be of use for studying nutrition and its related fields. In the current study, we examined how ChatGPT and Bing, which have been the most popular LLM-based generative AI models, respond to questions derived from the Japanese national registered dietitian licensure examination to evaluate the potential of the current LLM-based generative AI models as study aids for the students studying nutrition. Here, we demonstrate superiority of GPT-4-based Bing over ChatGPT based on GPT-3.5 in the correct answer rate. However, neither ChatGPT nor Bing can be considered an appropriate study aid at the moment, primarily due to their inconsistent responses and insufficient accuracy rates.

**Materials and Methods**

**Input datasets**

The input datasets are questions from the Japanese national registered dietitian licensure examination held in 2023, obtained from the official websites of the Ministry of Health, Labor and Welfare 2023). The questions consist of ten subject fields, which are A: Social Environment and Health, B: Human Body Structure, C: Food and Health, D: Basic Nutrition, E: Applied Nutrition, F: Nutrition Education, G: Clinical Nutrition, H: Public Nutrition, I: Food Service Management, and J: Application Problems. These questions were written in Japanese, and were directly utilized as they appeared for the datasets. Questions that included figures and/or graphs were excluded from the datasets (12 out of 200). All questions were input as a prompt in the independent sessions one by one.

**LLM-based generative AI models**

The version of ChatGPT used in this study was based on generative pre-trained transformer (GPT)-3.5, and this version of ChatGPT has been freely available through web browsers, which should make it one of the most popular LLM-based generative AI models. An updated version of the GPT, GPT-4 has been adopted by Bing and complimentarily available likewise. Bing possesses three different style models, ie., Precise, Balanced, and Creative. We compared these three styles and ChatGPT in the current study.

The GPT is a deep learning-based algorithm, and is a natural language processing model developed by OpenAI, GPT has been trained on vast amounts of text data, enabling it to generate coherent and contextually relevant responses to a wide range of queries or prompts, however, the training data in detail have not been publicly disclosed.

**Correct answer rates and response times**

Correct answer rates were determined based on the responses of each generative AI across ten independent trials for each question. The response times were manually determined by a stopwatch.

**Distribution of correct and incorrect answers**

Consistency of the responses was analyzed by evaluating the responses across 10 repeated attempts of the same three questions randomly selected from each subject field. Each response was produced by independent session, where the previous prompt did not influence the following results. To further evaluate the variability of responses, Shannon's entropy was calculated by negatively summing the product of the probability of each answer and logarithm (base 2) across all the five possible answers. The entropy was considered zero when all the ten answers were consistent. Precision Focus Index is defined as the product of the correct answer rate and (1-the entropy of incorrect answers normalized by maximum entropy) for each question.

**Evaluation of the effects of prompt strategies**

The effects of role and information assignment through prompt input on answer accuracy were evaluated using three approaches. First, we assessed the accuracy when the prompt "You are a highly experienced registered dietitian" was provided before answering each question. Additionally, we examined the effect of assigning a specialized role by specifying expertise in each subject area, such as "You are a highly experienced registered dietitian specializing in [subject area]" before answering questions in various professional fields. Lastly, we evaluated the responses when the AI models were provided with the exam questions, their correct answers, and brief explanations in advance before attempting to answer the questions.

**Ethical disclosure**

This study does not involve human subjects or identifiable private information.

**Statistical analyses**

The results are presented as mean ± standard error (S.E.). Comparisons between control group and each treatment group were analyzed by Dunnett's test. When comparing multiple groups, a parametric Tukey's or Tukey-Kramer's test was employed. A p-value of less than 0.05 was considered statistically significant.

## Results

**Overall correct answer rate**

Firstly, we evaluated the accuracy of each generative AI model in answering questions from the national licensure examination for registered dietitians to compare their overall performance (Fig. 1). Among these models, Bing-Precise demonstrated the highest overall correct answer rate (66.2±8.6%). Bing-Creative achieved a comparable score (61.4±4.9%), and both models exceeded the passing threshold of 60% in overall correct answer rate. In contrast, Bing-Balanced (43.3±4.2%) and ChatGPT (42.8±2.1%) showed lower correct answer rates and did not meet the passing criteria (Fig. 1).

**Correct Answer Rate by Subject Fields**

We further analyzed the performance of each model across ten subject fields relevant to the examination (Fig. 2). Except for Nutrition Education (field F), Bing-Precise and Bing-Creative consistently achieved correct answer rates that were 10–30% higher than those of Bing-Balanced and ChatGPT in nearly all subject fields. These results indicate that Bing-Precise and Bing-Creative generally perform better than other models in providing accurate answers across most areas of the examination. In addition, correct answer rates were generally high for all models in the fields of Social Environment and Health (field A), Human Body Structure (field B), and Basic Nutrition (field D), whereas all models showed relatively low correct answer rates in Nutrition Education (field F). In the field of Public Nutrition (field H), ChatGPT, which typically exhibited lower accuracy across most fields, outperformed Bing-Creative and achieved a correct answer rate comparable to that of Bing-Precise.

**Trends and Dispersion of Answers Over Repeated Attempts**

In our previous analyses, we noticed that even when the average correct answer rates appeared similar across independent sessions, the specific questions answered correctly often differed between sessions. In other words, the models frequently failed to provide consistent answers to the same questions across repeated attempts, and did not always reproduce the same (ideally correct) response.

Therefore, we aimed to investigate the degree of response consistency for each generative AI model. To this end, we randomly selected three questions from each subject field and asked each model to answer them ten times in independent sessions. The results are visualized in a heatmap-like format (Fig. 3A–D), where different patterns (such as white, black, stripes, or dots) represent different responses, allowing us to clearly observe the variability and consistency of the answers. To quantitatively evaluate the consistency and precision of the models' responses, we also calculated two metrics: Shannon's entropy (Fig. 3E) and the Precision Focus Index (Fig. 3F) for each question. Entropy measures the variability of responses across repeated attempts, with lower values indicating more consistent answers. The Precision Focus Index combines the correct answer rate with the consistency of incorrect responses, providing a comprehensive indicator of both accuracy and stability.

Bing-Precise and Bing-Creative demonstrated relatively stable performance, with a high proportion of questions answered correctly in all attempts (Fig. 3A, C, E). ChatGPT exhibited lower variability in responses despite its relatively low correct answer rate (Fig. 3D, E). In contrast, Bing-Balanced exhibited greater variability, frequently providing inconsistent answers to the same questions (Fig. 3B, E). The entropy analysis clearly demonstrated the relative instability of Bing-Balanced (0.84±0.10, 1.33±0.10, 0.67±0.12, 0.68±0.13 for Bing-Precise, -Balanced, -Creative, ChatGPT, respectively; maximum possible value is 2.32) (Fig. 3E). The Precision Focus Index was lowest for Bing-Balanced (0.29±0.05; although not statistically significant), while the other three models showed higher but still moderate values (0.50±0.06, 0.43±0.08, 0.48±0.07 for Bing-Precise, -Creative, ChatGPT, respectively), indicating that none of the models achieved high consistency or precision (Fig. 3F). Further analysis of the number of questions answered correctly across all ten attempts revealed that, while Bing-Precise achieved the highest total number of correct answers, the proportion of questions answered correctly in all ten sessions was as small as 10% (Fig. 3G), indicating that Bing-Precise demonstrated limited consistency in providing the correct answer across repeated attempts even with high overall accuracy. In contrast, both Bing-Creative and ChatGPT showed a relatively higher probability of providing the same answer-either correct or incorrect-across all sessions. This may indicate greater consistency in response patterns, regardless of their correctness (Fig. 3G). Bing-Balanced, on the other hand, exhibited a lower total number of correct answers and a lower probability of consistently providing the same answer, whether correct or incorrect, compared to the other models (Fig. 3G).

**Average Response Time for Answering the Questions**

We also measured the average response time required by each model to answer the examination questions (Fig. 4). ChatGPT responded most rapidly (13.8±2.3 sec.), followed by Bing-Balanced (23.2±2.9 sec.), while Bing-Precise (55.8±7.4 sec.) and Bing-Creative (71.0±15.3 sec.) required a longer average response time.

**Effects of Prompt Strategies on ChatGPT's Answer Accuracy**

Finally, we investigated the impact of various prompt strategies on the accuracy of ChatGPT's answers (Fig. 5). The use of role assignment prompts such as "You are a highly experienced registered dietitian" and field-specific expertise prompts like "You are a highly experienced registered dietitian specializing in the corresponding subject area (Field A–I)" did not improve the correct answer rate, either overall or within any individual subject field. However, a significant increase in the correct answer rate was observed only when ChatGPT was provided with the correct answers and brief explanations as part of the prompt. Nevertheless, even after explicitly providing the correct answers, the overall correct answer rate improved by only about 10% (42.8±0.7% to 53.5±0.7%). This improvement appeared only in certain subject fields, specifically Fields A, B, C, G, and I (Fig. 5).

**Discusstion**

To the best of our knowledge, this is the first study to evaluate the potential of LLM-based generative AI models as assistant tools for students enrolled in registered dietitian courses, particularly in the context of the Japanese national licensure examination for registered dietitians.

Although some LLM-based generative AI models achieved overall correct answer rates exceeding the passing threshold for the national registered dietitian licensure examination, several critical issues were identified in this study (Fig. 1). Notably, even the best-performing models demonstrated limited consistency across repeated attempts, with only 10-30% of questions answered correctly in all sessions (Fig. 3G). Furthermore, prompt engineering strategies, such as role assignment and field-specific expertise, failed to produce any significant improvement in answer accuracy; significant gains were observed only when correct answers and explanations were explicitly provided in the prompts (Fig. 5). These findings highlight the current limitations of using LLM-based generative AI as a reliable support tool for students in nutrition and dietitian education.

In particular, our study is the first to highlight the critical issue of answer consistency in the context of learning for medical and health-related fields, specifically in the training of registered dietitians (Fig. 3). While previous studies have primarily focused on the accuracy and overall performance of LLM-based generative AI models, our results highlight that consistency of responses, that is, the capacity of an AI to provide stable and reproducible answers to the same questions across repeated attempts, is an essential prerequisite for their practical application as educational tools in professional training settings. Inconsistent AI-generated answers may lead to confusion or misunderstanding among students, potentially undermining the acquisition of accurate and trustworthy knowledge that is important for future healthcare professionals. Our results suggest that, for LLM-based generative AI models to be successfully integrated into registered dietitian education, future development should prioritize not only accuracy but also the reproducibility of AI outputs.

In the context of LLM-based generative AI, the consistency and variability of generated responses are primarily determined by a parameter known as "temperature." The temperature parameter is a value that controls the randomness of the model's output. Lower temperature values (close to 0) make the model's responses more deterministic, often resulting in more consistent and predictable answers (Ackley et al., 1985; Peeperkorn , 2024; Roemmele & Gordon, 2018). In contrast, higher temperature values increase randomness, allowing for more diverse and creative outputs, but potentially reducing consistency. Therefore, for the purpose of educational support, it is essential that the temperature parameter is set close to zero, ensuring consistency and reproducibility in AI-generated responses. Although the temperature values used in the general-purpose generative AI models such as ChatGPT and Bing are not publicly disclosed, previous reports and official documentation suggest that default values are typically set around 0.7 to 0.8, which favors more diverse and creative outputs rather than deterministic ones (Ming et al., 2024). Our results indicate that, for all models tested, the temperature settings are likely too high for educational support purposes, as evidenced by the lack of consistent answers across repeated attempts (Fig. 3). Since users generally cannot freely adjust the temperature parameter in these proprietary generative AI models, the use of dedicated internal LLM environments or API-based solutions, where temperature can be explicitly set to a sufficiently low value is recommended for educational applications.

Of course, high consistency does not necessarily imply high accuracy. Therefore, in addition to evaluating entropy, we adopted the Precision Focus Index to simultaneously assess both the stability and correctness of the responses (Fig. 3F). Although Bing-Precise, Bing-Creative, and ChatGPT demonstrated higher Precision Focus Index values compared to Bing-Balanced, none of these models exceeded a value of 0.5, indicating that they were unable to provide consistently correct answers across repeated attempts. Thus, even the better-performing models were suggested to fall short of achieving the level of stability and accuracy required for reliable educational support.

Several evaluations related to the preparation for licensure exams have already been conducted in the fields of medicine, nursing and pharmacy, in which LLMs have demonstrated promising, though not perfect, performance on national licensure examinations. For example, studies have shown that models such as GPT-4 can achieve correct answer rates exceeding the passing threshold on the Japanese National Medical Licensure Examination and the National Pharmacist Examination (Kataoka et al., 2023; Kunitsu, 2023). These findings indicate the potential of LLMs as support tools for medical and pharmacy students. However, these studies also highlight ongoing limitations, such as variability in performance depending on the field and the type of questions, as well as persistent challenges regarding consistency and reasoning. In particular, recent studies have reported that, while LLMs could attain high accuracy in certain medical and nursing domains, their performance varied across specialties and question formats (Haze et al., 2023; Kaneda et al., 2023; Meyer et al., 2024; Ming et al., 2024). Similarly, Kataoka et al. and Kunitsu reported that, although LLMs surpassed the pass mark in some areas of the Japanese medical and pharmacist licensure exams, their responses were not consistently reliable or fully aligned with expert-level reasoning (Kataoka et al., 2023; Kunitsu, 2023). Similar findings have been observed in nursing and non-Japanese examinations; for example, Taira et al. showed that ChatGPT met or nearly met the passing threshold on recent Japanese National Nurse Examinations (Taira et al., 2023). Likewise, Wang et al. found that ChatGPT-3.5 came close to passing but ultimately failed the Taiwanese Pharmacist Licensing Examination, with performance varying across subjects and languages (Ming et al., 2024; Wang et al., 2023). Our observations revealed that Bing-Precise and Bing-Creative, which are based on GPT-4, showed superior performance to GPT-3.5-based ChatGPT in terms of correct answer rate in the field of nutrition and dietetics. This finding is consistent with previous reports demonstrating the superiority of GPT-4 over GPT-3.5 in other medical-related fields (Haze et al., 2023; Kaneda et al., 2023; Wang et al., 2023; Yeadon et al., 2024).

Although strict comparisons across different licensure examinations are inherently limited, our findings indicate that the performance of generative AI models on the registered dietitian examination was generally lower than what has been reported for medical and pharmacy licensure tests. We suggest that this discrepancy should be attributed not only to limitations in the reasoning capabilities of GPT models but also to biases in the pre-training data. Compared to medicine, information related specifically to nutrition and dietetics is less abundant and less frequently represented in academic literature and public online resources. As a result, proprietary general-purpose LLMs are likely exposed to a much broader and deeper corpus of general medical knowledge during pre-training, while their coverage of dietitian-specific content remains relatively sparse (Haze et al., 2023). Accordingly, these models appear to be better equipped to address questions rooted in general medicine than those requiring specialized expertise in nutrition and dietetics. This is further supported by our field-specific analysis, which showed higher correct answer rates in subject areas related to basic medical sciences including human body structure and basic nutrition, but notably lower performance in fields such as nutrition education, where available information is more limited and specialized (Fig. 2). Therefore, these models will require intensive and targeted fine-tuning with nutrition- and dietitian-focused resources, particularly those addressing more specialized topics, to enhance their usefulness for registered dietitian education.

Another important aspect revealed in this study is the limited effectiveness of prompt engineering strategies in improving answer accuracy (Fig. 5). While prompt engineering has been widely discussed as a means to enhance LLM performance (Schulhoff et al., 2025), our findings suggest that, at least in the context of the Japanese registered dietitian licensure examination, simply assigning professional roles or specifying field expertise does not meaningfully improve the correct answer rate. Prompt engineering may sometimes lead LLMs to more relevant and contextually appropriate responses, but it is unable to compensate for fundamental gaps in domain knowledge or reasoning ability inherent to the model's training data and architecture.

The observed trade-off between response speed and answer accuracy is also noteworthy (Fig. 4). While ChatGPT provided the fastest responses, it was less accurate than Bing-Precise and Bing-Creative, which took significantly longer to generate answers. These findings may reflect differences in model architecture, search strategies, or resource allocation. This trade-off between speed and accuracy should be carefully considered when deploying AI models. In contexts such as exam preparation, balancing these factors will be essential to maximize the utility of AI-assisted tools.

In this study, we evaluated the performance of generative AI models using the original Japanese-language questions from the national examination conducted in Japan. It is well known that the choice of language can significantly affect the performance of generative AI, and results obtained in Japanese may differ from those evaluated in English, which is often considered the standard for such assessments (Wang et al., 2023; Zhong et al., 2024). To preliminarily assess the potential impact of language, we translated the exam questions into English using DeepL (DeepL SE, Cologne, Germany), a leading machine translation AI, and used these English versions as prompts. The results showed that, at least in our preliminary analysis, translation had little effect on the overall accuracy rates. Given that it is more natural for the students in Japan to use study aids in Japanese, we conducted the main evaluation using the original Japanese questions. Future enhancements in the multilingual capability of generative AI are anticipated to improve answer quality in non-English languages.

This study was conducted using ChatGPT (GPT-3.5) and Bing (GPT-4) as chat models available as of 2024. Given the rapid advancements in generative AI, newer multimodal models such as GPT-4o, which are now implemented in the current versions of ChatGPT and Copilot, may demonstrate improved performance—particularly in terms of accuracy, consistency, and multilingual capabilities. Future studies should therefore evaluate these latest technologies.

**Conclusion**

This study provides the first comprehensive evaluation of LLM-based generative AI as a support tool for students preparing for the Japanese registered dietitian licensure examination. While some models achieved correct answer rates above the passing threshold, we identified several critical limitations that must be addressed before these technologies can be reliably integrated into professional education. Most notably, the lack of answer consistency across repeated attempts and the limited effectiveness of prompt engineering strategies highlight the need for improvements in both model architecture and parameter control. Furthermore, current LLMs possess insufficient domain-specific knowledge in nutrition and dietetics, likely due to biases in pre-training data.

Overall, while LLM-based generative AI holds promise as a supplementary educational resource, our findings indicate that, as in the field of nutrition and dietetics, careful consideration of their current strengths and limitations is required when applying LLMs in health professions education, including for registered dietitians.

**Figure legends**

Fig.1

Overall correct answer rate

The overall correct answer rates for 200 questions from the Japanese national registered dietitian licensure examination produced by Bing-Precise (open bar), Bing-Balanced (hatched bar), Bing-Creative (dotted bar), and ChatGPT (closed bar). Data are presented as mean±S.E., **; $p<0.01$, n=10.

Fig. 2

Correct answer rate by subject fields

The correct answer rates for each of the ten subject fields. The results were generated by Bing-Precise (open bar), Bing-Balanced (hatched bar), Bing-Creative (dotted bar), and ChatGPT (closed bar). The subject fields are A: Social Environment and Health, B: Human Body Structure, C: Food and Health, D: Basic Nutrition, E: Applied Nutrition, F: Nutrition Education, G: Clinical Nutrition, H: Public Nutrition, I: Food Service Management, and J: Application Problems. Data are presented as mean±S.E., a; $p<0.05$, aa; $p<0.01$ vs Bing-Precise, b; $p<0.05$, bb; $p<0.01$ vs Bing-Balanced, c; $p<0.05$, cc; $p<0.01$ vs Bing-Creative, d; $p<0.05$, dd; $p<0.01$ vs ChatGPT, n=10.

Fig. 3

Trends and dispersion of answers over repeated attempts

Distribution of correct and incorrect responses across 10 repeated attempts in Bing-Precise(A), Bing-Balanced (B), Bing-Creative (C), and ChatGPT (D). Three questions from each subject field were randomly selected for the analysis. White cells represent correct answers, while different patterns indicate distinct incorrect answers. Results are shown for 10 repeated attempts on the same questions. Comparison of response entropy (E) and Precision Focus Index (F) for each model (F). Bing-Precise (open bar), Bing-Balanced (hatched bar), Bing-Creative (dotted bar), and ChatGPT (closed bar). Data are presented as mean±S.E., **; $p<0.01$, n=30, Distribution of questions based on the number of correct answers achieved over 10 repeated attempts (G). White cells indicate questions answered correctly in all 10 attempts, while black cells show consistent incorrect responses. Gradation from white to black represents the proportion of correct answers across the 30 questions.

Fig. 4

Average response time for answering the questions

The average response time required for answering each question in Bing-Precise (open bar), Bing-Balanced (hatched bar), Bing-Creative (dotted bar), and ChatGPT (closed bar). Data are presented as mean±S.E., **; $p<0.01$, n=30.

Fig. 5

Effects of prompt strategies on ChatGPT's answer accuracy

The effects of role assignment, field expertise, and contextual knowledge provided through prompts on the correct answer rates of ChatGPT. The correct answer rates are shown for responses without prompts (open bar), after the prompt, "You are a highly experienced registered dietitian." (hatched bar), and after the field-specific prompt, "You are a highly experienced registered dietitian specializing in the corresponding subject area (A-I)." The correct answer rates are also shown after providing correct answers with brief explanations (closed bar). The subject fields are A: Social Environment and Health, B: Human Body Structure, C: Food and Health, D: Basic Nutrition, E: Applied Nutrition, F: Nutrition Education, G: Clinical Nutrition, H: Public Nutrition, and I: Food Service Management. Data are presented as mean±S.E., *; $p<0.05$, **; $p<0.01$, n=10.

**Statements and Declarations**

**Funding**

This study was partly supported by Information Education Research Center, Kanagawa Institute of Technology.

**Competing interests**

The authors declare no conflict of interest related to the current study.

**Author contributions**

*Yuta Nagamori, Mikoto Kosai, Yuji Kawai, and Licht Miyamoto mainly performed data acquisition, and prepared figures. Licht Miyamoto conceived, designed, and conducted the study, and wrote the manuscript. Yuta Nagamori and Mikoto Kosai also contributed to the study design. The others contributed to the data acquisition and/or interpretation.*

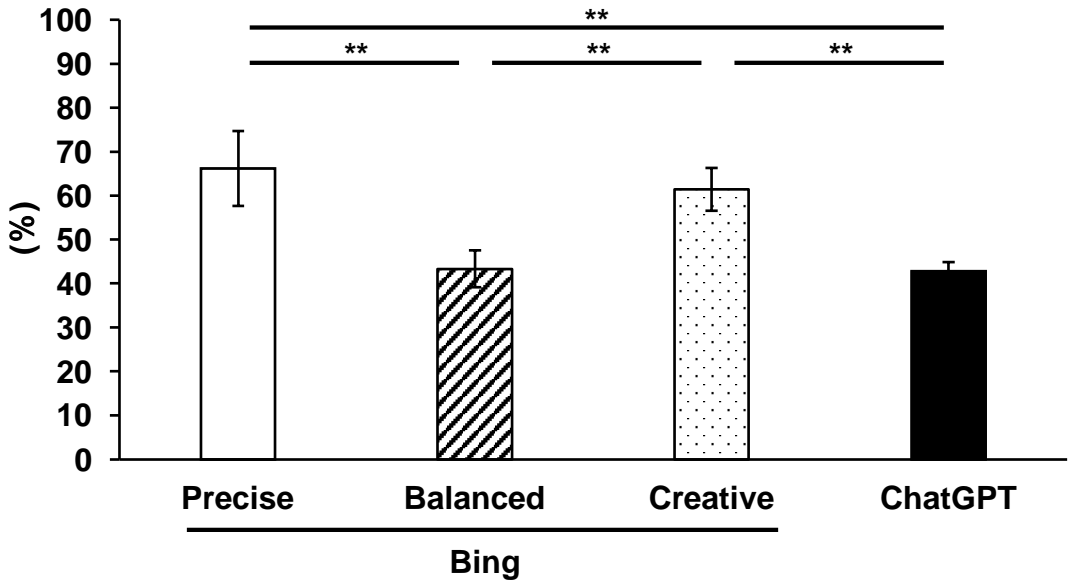

Fig. 1

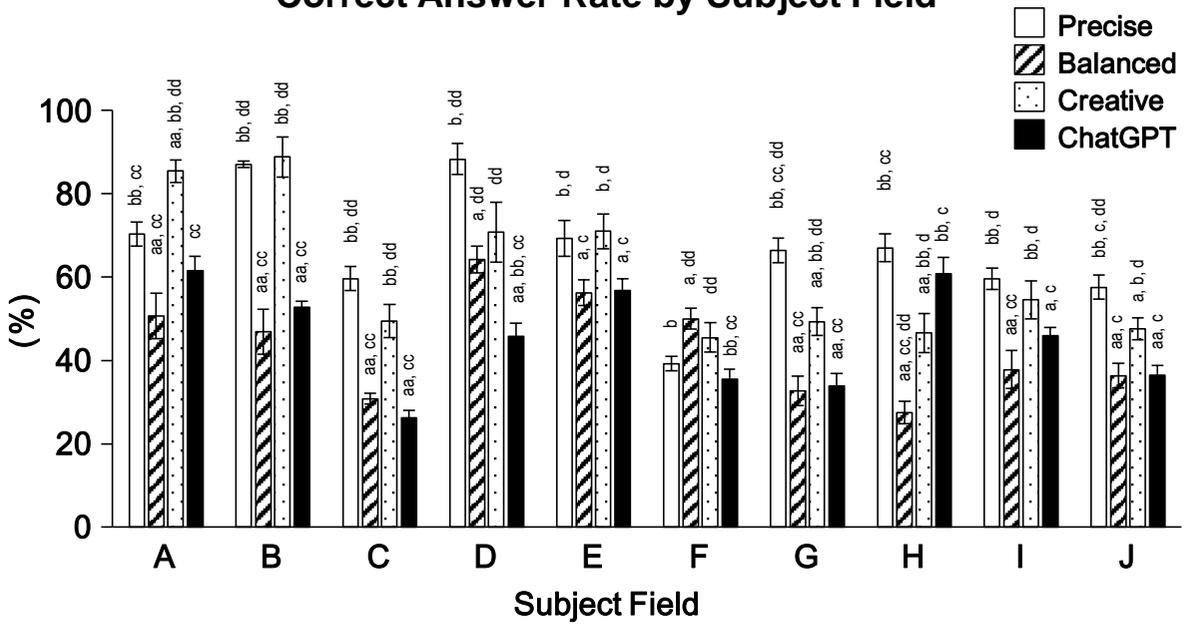

Fig. 2

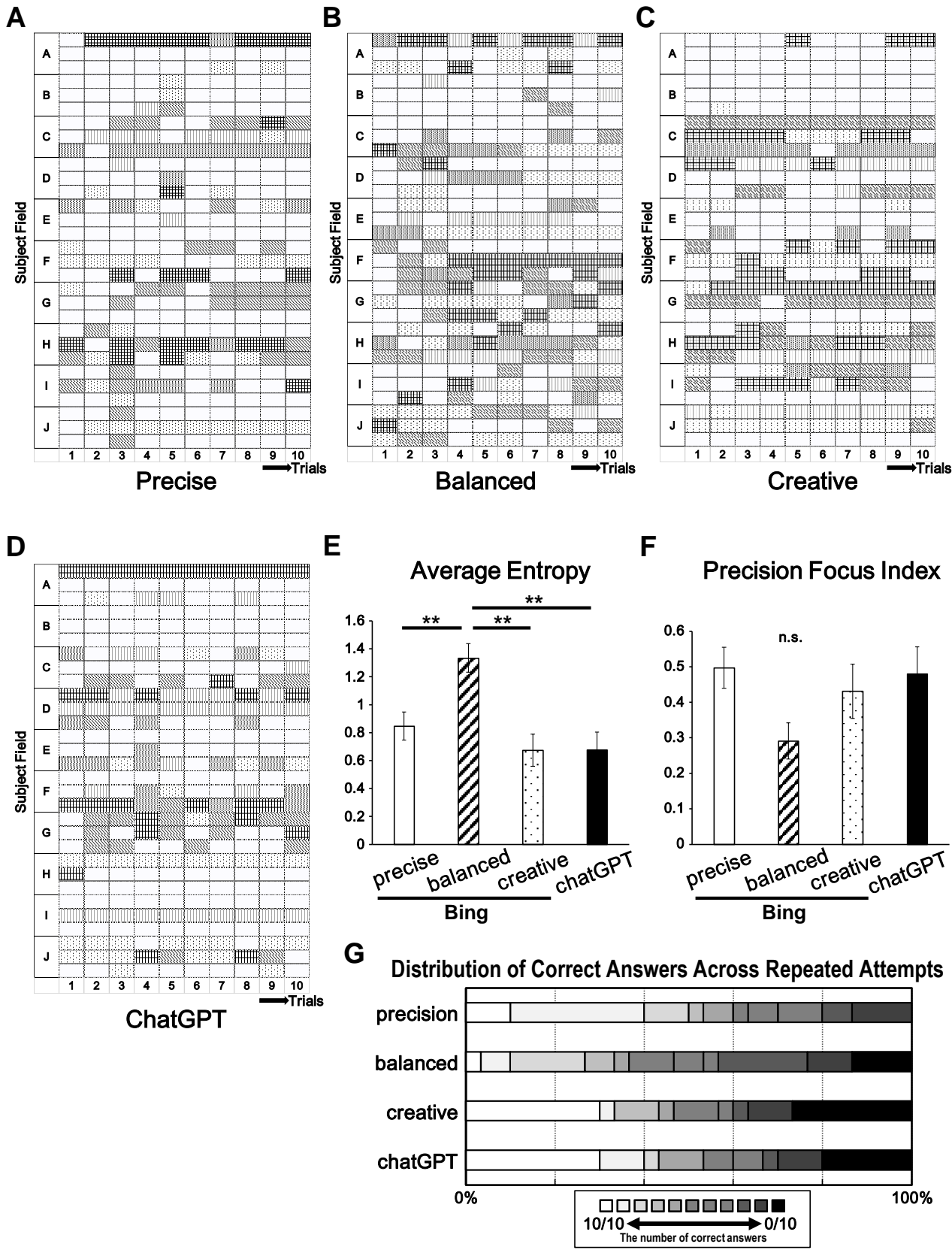

Fig. 3

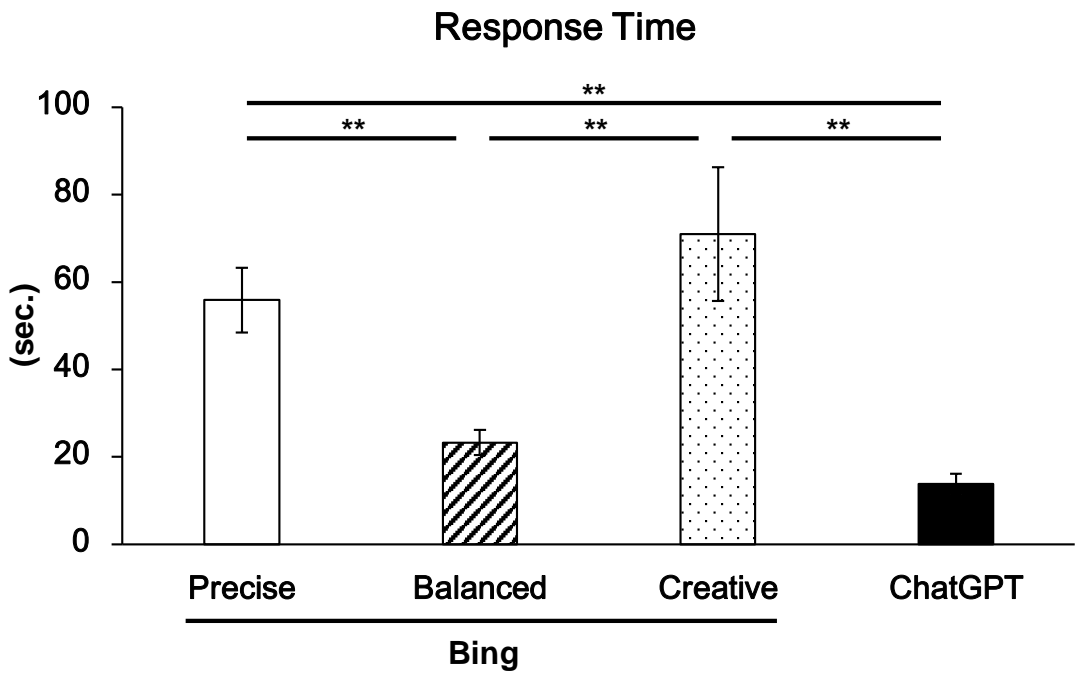

Fig. 4

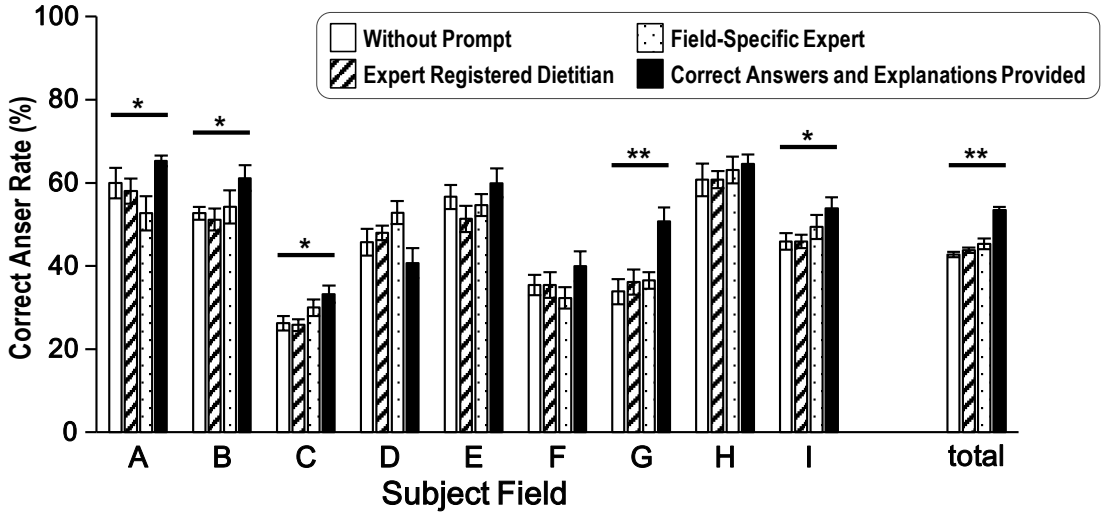

Fig. 5